\documentclass[11pt,letterpaper]{article}
\usepackage{emnlp2017}
\usepackage{times}
\usepackage{latexsym}
\usepackage{todonotes}
\usepackage{algorithm}
\usepackage{algpseudocode}

\emnlpfinalcopy

\def\emnlppaperid{***}

\usepackage{enumitem}  
\usepackage{url}
\urldef\pubmedquery\url{https://www.ncbi.nlm.nih.gov/pubmed?term=(%222016%22%5BDate%20-%20Publication%5D%20%3A%20%222016%22%5BDate%20-%20Publication%5D)}

\title{Learning what to read: Focused machine reading}  

\author{Enrique Noriega-Atala \\ {\bf Marco A. Valenzuela-Esc\'{a}rcega}  \And Clayton T. Morrison \\  {\bf Mihai Surdeanu} \AND {\normalfont University of Arizona}\\ Tucson, Arizona, USA \\ \\ {\tt \{enoriega,marcov,claytonm,msurdeanu\}@email.arizona.edu} }

\date{}

\begin{document}

\maketitle

\begin{abstract}
Recent efforts in bioinformatics have achieved tremendous progress in the machine reading of biomedical literature, and the assembly of the extracted biochemical interactions into large-scale models such as protein signaling pathways. 
However, batch machine reading of literature at today's scale (PubMed alone indexes over 1 million papers per year) is unfeasible due to both cost and processing overhead. In this work,
we introduce a focused reading approach to guide the machine reading of biomedical literature towards {\em what} literature should be read to answer a biomedical query as efficiently as possible. We introduce a family of algorithms for focused reading, including an intuitive, strong baseline, and a second approach which uses a reinforcement learning (RL) framework that learns when to explore (widen the search) or exploit (narrow it). 
We demonstrate that the RL approach is capable of answering more queries than the  baseline, while being more efficient, i.e., reading fewer documents.

\end{abstract}

\section{Introduction}

The millions of academic papers in the biomedical domain contain a vast amount of information that may lead to new hypotheses for disease treatment.  However, scientists are faced with a problem of ``undiscovered public knowledge,'' as they struggle to read and assimilate all of this information~\cite{swanson1986undiscovered}.  Furthermore, the literature is growing at an exponential rate~\citep{pautasso2012}; PubMed\footnote{\url{http://www.ncbi.nlm.nih.gov/pubmed}} has been adding more than a million papers per year since 2011. 
We have surpassed our ability to keep up with and integrate these findings through manual reading alone.

Large ongoing efforts, such as the BioNLP task community \cite{nedellec2013overview,kim2012genia,kim2009overview} and the DARPA Big Mechanism Program \cite{cohen2015}, are making progress in advancing methods for machine reading and assembly of extracted biochemical interactions into large-scale models.  However, to date, these methods rely either on the manual selection of relevant documents, or on the processing of large batches of documents that may or may not be relevant to the model being constructed.  

Batch machine reading of literature at this scale poses a new, growing set of problems.  
First, access to some documents is costly.  The PubMedCentral (PMC) Open Access Subset\footnote{\url{https://www.ncbi.nlm.nih.gov/pmc/tools/openftlist/}} (OA) is estimated\footnote{\url{https://tinyurl.com/bachman-oa}} to comprise 20\%\footnote{This includes 5\% from PMC author manuscripts.} of the total literature; the remaining full-text documents are only available through paid access.   
Second, while there have been great advances in quality, machine reading is still not solved.  Updates to our readers requires reprocessing the documents.  For large document corpora, this quickly becomes the chief bottleneck in information extraction for model construction and analysis.
Finally, even if we could cache all reading results, the search for connections between concepts within the extracted results should not be done blindly.  At least in the biology domain, the many connections between biological entities and processes leads to a very high branching factor, making blind search for paths intractable.

To effectively read at this scale, we need to incorporate methods for {\em focused reading}: develop the ability to pose queries about concepts of interest and perform targeted, incremental search through the literature for connections between concepts while minimizing reading documents that are likely irrelevant.

In this paper we present what we believe is the first algorithm for focused reading.  We make the following contributions:\\
{\noindent {\bf (1)}} Present a general framework for a family of possible focused reading algorithms along with a baseline instance.\\
{\noindent {\bf (2)}} Cast the design of focused reading algorithms in a reinforcement learning (RL) setting, where the machine decides if it should explore (i.e., cast a wider net) or exploit (i.e., focus reading on a specific topic).\\
{\noindent {\bf (3)}} Evaluate our focused reading policies in terms of search efficiency and quality of information extracted. The evaluation demonstrates the effectiveness of the RL method: this approach found more information than the strong baseline we propose, while reading fewer documents. 
\section{Related Work}\label{sec:related}


The past few years have seen a large body of work on information extraction (IE), particularly in the biomedical domain. 
This work is too vast to be comprehensively discussed here. We refer the interested reader to the BioNLP community~\cite[inter alia]{nedellec2013overview,kim2012genia,kim2009overview} for a starting point. 
However, most of this work focuses on {\em how} to read, not on {\em what} to read given a goal. To our knowledge, we are the first to focus on the latter task.

Reinforcement learning has been used to achieve state of the art performance in several natural language processing (NLP) and information retrieval (IR) tasks.
For example, RL has been used to guide IR and filter irrelevant web content~\cite{seo2000reinforcement, zhang2001personalized}.
More recently, RL has been combined with deep learning with great success, e.g., for improving coreference resolution~\cite{clark2016deep}. 
Finally, RL has been used to improve the efficiency of IE by learning how to incrementally reconcile new information and help choose what to look for next \cite{narasimhan2016improving}, a task close to ours.
This serves as an inspiration for the work we present here, but with a critical difference: \citet{narasimhan2016improving} focus on slot filling using a pre-existing template. This makes both the information integration and stopping criteria well-defined. On the other hand, in our focused reading domain, we do not know ahead of time which new pieces of information are necessarily relevant and must be taken in context.
\section{Focused Reading}\label{sec:focusedreading}

\begin{figure}[t]
\begin{center}
  \includegraphics[width=.4\textwidth]{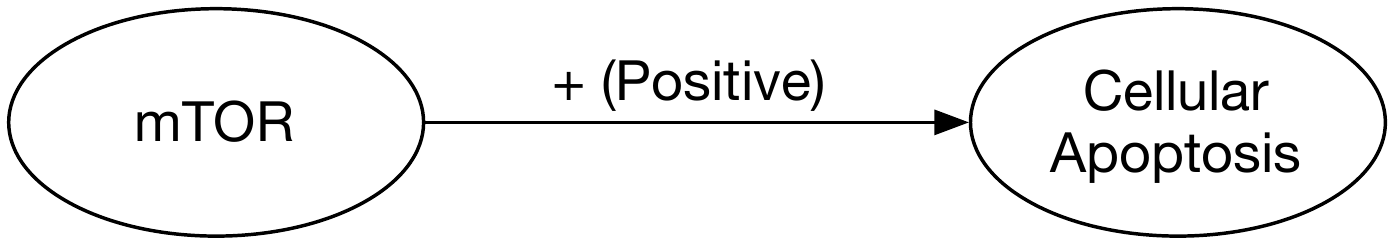}
  \caption{Example of a graph edge encoding the relation extracted from the text: \emph{mTOR triggers cellular apoptosis}.}\label{fig:fragment}
  \vspace{-3mm}
  \end{center}
\end{figure}

Here we consider focused reading for the biomedical domain, and we focus on binary promotion/inhibition interactions between biochemical entities. 
In this setting, the machine reading (or IE) component constructs a directed graph, where vertices represent {\em participants} in an interaction (e.g., protein, gene, or a biological process), and edges represent directed activation interactions. 
Edge labels indicate whether the controller entity has a {\em positive} (promoting) or {\em negative} (inhibitory) influence on the controlled participant. 
Figure~\ref{fig:fragment} shows an example edge in this graph.


We use REACH\footnote{\url{https://github.com/clulab/reach}}, an open source IE system~\cite{Valenzuela:15}, to extract interactions from unstructured biomedical text and construct the graph above. 
We couple this IE system with a Lucene\footnote{\url{https://lucene.apache.org}} index of biomedical publications to retrieve papers based on queries about participant mentions in the text (as discussed below).

Importantly, we essentially use IE as a black box (thus, our method could potentially work with any IE system), and focus on strategies that guide {\em what} the IE system reads for a complex information need.
In particular, we consider the common scenario where a biologist (or other model-building process) queries the literature on:
\begin{quote}
 How does one participant (source) affect another (destination), where the connection is typically indirect?
 \end{quote}
 This type of queries is common in biology, where such direct/indirect interactions are observed in experiments, but the explanation of why these dependencies exist is unclear.
 
Algorithm~\ref{alg:focusedreading} outlines the general focused reading algorithm for this task.  
In the algorithm, $S, D, A$, and $B$ represent individual participants, where $S$ and $D$ are the {\bf s}ource and {\bf d}estination entities in the initial user query. $G$ is the interaction graph that is iteratively constructed during the focused reading procedure, with $V$ being the set of vertices (biochemical entities), and $E$ the set of edges (promotion/inhibition interactions). 
$\mathit{\Sigma}$ is the strategy that chooses which two entities/vertices to be used in the next information retrieval iteration.
$Q$ is a Lucene query automatically constructed in each iteration to retrieve new papers to read.


\begin{algorithm}
\caption{Focused reading framework}
\begin{algorithmic}[1]
{\small
\Procedure{FocusedReading}{$S,D$}
   \State {\scriptsize $G \gets \{\{S, D\}, \emptyset\}$} \label{alg:fr:graph}
   \Repeat \label{alg:fr:start}
   		\State {\scriptsize $\mathit{\Sigma} \gets $ \Call{EndpointStrategy}{$G$}} \label{alg:endpointsstrategy}
		\State {\scriptsize $(A,B) \gets $ \Call{ChooseEndPoints}{$\mathit{\Sigma},G$}} \label{alg:chendpoints}
   		\State {\scriptsize $Q \gets $ \Call{ChooseQuery}{$A,B,G$}} \label{alg:chquery}
   		\State {\scriptsize $(V, E) \gets$ \Call{Lucene+Reach}{$Q$}} \label{alg:ie}
   		\State {\scriptsize \Call{Expand}{$V, E, G$}} \label{alg:reconcile}
   \Until{{\scriptsize \Call{IsConnected}{$S,D$}} OR {\scriptsize \Call{StopConditionMet}{$G$}}}\label{alg:fr:stop}
\EndProcedure
}
\end{algorithmic}
\label{alg:focusedreading}
\end{algorithm}

The algorithm initializes the search graph as containing the two unconnected participants as vertices: $\{S, D\}$ (line~\ref{alg:fr:graph}).
The algorithm then enters into its central loop (lines \ref{alg:fr:start} through \ref{alg:fr:stop}). The loop terminates when one or more directed paths connecting $S$ to $D$ are found, or when a stopping condition is met: either $G$ has not changed since the previous run through the loop, or after exceeding some number of iterations through the loop (in this work, ten).

At each pass through the loop the algorithm grows the search graph as follows:
\setlist{nolistsep}

\begin{enumerate}
\item The graph $G$ is initialized with two nodes, the source and destination in the user's information need, and no edges (because we have not read any papers yet).
  \item Given the current graph, choose a strategy, $\mathit{\Sigma}$, for selecting which entities to query next: {\em exploration} or {\em exploitation} (line~\ref{alg:endpointsstrategy}). In general, exploration aims to widen the search space by adding many more nodes to the graph, whereas exploitation aims to narrow the search by focusing on entities in a specific region of the graph.
  \item Using strategy $\mathit{\Sigma}$, choose the next entities to attempt to link: $(A,B)$ (line~\ref{alg:chendpoints}). 
  \item Choose a query, $Q$: again, using {\em exploration} or {\em exploitation}, following the same intuition as with the entity choice strategy (line~\ref{alg:chquery}).
Here exploration queries retrieve a wider range of documents, while exploitation queries are more restrictive.
  
  \item Run the Lucene query to retrieve papers and process the papers using the IE system.  The result of this call is a set of interactions, similar to that in Figure~\ref{fig:fragment} (line~\ref{alg:ie}).
  \item Add the new interaction participant entities (vertices $V$) and directed influences (edges $E$) to the search graph (line~\ref{alg:reconcile}).
    \item If the source and destination entities are connected in $G$, stop: the user's information need has been addressed. Otherwise, continue from step 2.
\end{enumerate}

The central loop performs a bidirectional search in which each iteration expands the search horizon outward from $S$ and $D$.  
Algorithm~\ref{alg:focusedreading} represents a family of possible focused reading algorithms, differentiated by how each of the functions in the main loop are implemented.  In this work, {\sc IsConnected} stops after a single path is found, but a variant could consider finding multiple paths, paths of some length, or incorporate other criteria about the properties of the path.  We next consider particular choices for the inner loop functions.
\section{Baseline Algorithm and Evaluation}\label{sec:dataset}

\begin{table*}[t!]
  \begin{center}
  {\footnotesize
  \begin{tabular}{r|c|cc}
  	& \textbf{Baseline} & \textbf{RL Query Policy} & \\
  	\hline
    \textbf{\# IR queries} &  573 & 433 & 25\% decrease\\
    \textbf{Unique papers read} & 26,197 & 19,883 & 24\% decrease\\
    \textbf{\# Paths recovered} (out of 289) & 189 (65\%) & 198 (68\%) & 3\% increase
  \end{tabular}
  \caption{Results of the baseline and RL Query Policy for the focused reading of biomedical literature.}
  \vspace{-6mm}
  \label{table:reading-results}
  }
  \end{center}
\end{table*}

The main functions that affect the search behavior of Algorithm~\ref{alg:focusedreading} are {\sc EndpointStrategy} and {\sc ChooseQuery}.
Here we describe a {\em baseline} focused reading implementation in which {\sc EndpointStrategy} and {\sc ChooseQuery} aim to find any path between $S$ and $D$ as quickly as possible.

For {\sc EndpointStrategy}, we follow the intuition that some participants in a biological graph tend to be connected to more participants than others, and therefore more likely to yield interactions providing paths between participants in general.  Our heuristic is therefore to choose new participants to query that currently have the most inward and outgoing edges (i.e., highest vertex degree) in the current state of $G$ (disallowing choosing an entity pair used in a previous query).

Now that we have our candidate participants $(A,B)$, our next step is to formulate how we will use these participants to retrieve new papers.  Here we consider two classes of query: (1) we restrict our query to only retrieve papers that simultaneously mention both $A$ and $B$, therefore more likely retrieving a paper with a direct link between $A$ and $B$ ({\em exploit}), or (2) we retrieve papers that mention either $A$ or $B$, therefore generally retrieving more papers that will introduce more new participants ({\em explore}).  For our baseline, where we are trying to find a path between $S$ and $D$ as quickly as possible, we implement a greedy {\sc ChooseQuery}: first try the conjunctive exploitation query; if no documents are retrieved, then ``relax'' the search to the disjunctive exploration query.



To evaluate the baseline, we constructed a data set based on a collection of papers seeded by a set of 132 entities that come from the University of Pittsburgh DyCE\footnote{{\bf Dy}namic {\bf C}ell {\bf E}nvironment model of pancreatic cancer.} model, a biomolecular model of pancreatic cancer~\cite{dyce-inprep}.  Using these entities, we retrieved 70,719 papers that mention them.  We processed all papers using REACH, extracting all of the interactions mentioned, and converted them into a single graph.
The resulting graph consisted of approximately 80,000 vertices, 115,000 edges, and had an average (undirected) vertex degree of 24.  We will refer to this graph as the {\em REACH graph}, as it represents what {\em can} be retrieved by REACH from the set of 70K papers.  Next, we identified which pairs of the original 132 entities are connected by directed paths in DyCE.  A total of 789 pairs were found.  We used 289 of these entity pairs as testing queries (i.e., generating queries that aim to explain how a given pair is connected according to the literature). The other 500 pairs were held out to train the RL method described below.

We ran this baseline focused reading algorithm on each of the 289 pairs of participants, in each case attempting to recover a directed path from one to the other.  The results are summarized in the middle column of Table~\ref{table:reading-results}.  By issuing 573 queries, the baseline read 26,197 papers out of the total 70,719 papers (37\% of the corpus), in order to recover 189 of the 289 paths (65\%).

\section{Reinforcement Learning for Focussed Reading}\label{sec:rl}

We analyzed the baseline's behavior in the evaluation to identify the conditions under which it failed to find paths.  From this, we found that some of the failures could be avoided had we used a different strategy for {\sc ChooseQuery}, i.e., the baseline chose to exploit when it should have explored more.  The conditions for making different choices depend on the current state of $G$, and earlier query behavior can affect later query opportunities, making this an iterative decision making problem and a natural fit for a RL formulation.

Inspired by this observation, we consider RL for finding a better policy for {\sc ChooseQuery}.  We'll refer to an instance of the focused reading algorithm with a learned {\sc ChooseQuery} policy as the {\em RL Query Policy}.  All other focus reading functionality is the same as in the baseline.  
For actions, we consider a simple binary action choice: exploit (conjunctive query) or explore (disjunctive query).
We represent the state of the search using a set of features that include: 
(f1) the current iteration of the search; 
(f2) the number of times a participant has been used in previous queries;
(f3) whether the participants are chosen from the same connected component in $G$;
(f4) the vertex degree of participants; 
and (f5) the search iteration in which a participant was introduced.
With the goal of recovering paths as quickly as possible, we provide a reward of $+1$ if the algorithm successfully finds a path, a reward of $-1$ if the search fails to find a path, and assess a ``living reward'' of $-0.05$ for each step during the search, to encourage trying to finish the search as quickly as possible.

\begin{table*}[t!]
\begin{footnotesize}
\begin{center}
  \begin{tabular}{r|cccccc}
    & All & $-$ Iteration & $-$ Query& $-$ Same & $-$ Ranks (f4)& $-$ Particip.\\
    & features & number (f1) & counts (f2)& component (f3)& & intro. (f5)\\
    \hline
    \emph{Paths found}& 198&199 & 200 & 201 & \textbf{202} & 196\\
    \emph{Papers read}& 19,883&20,918 & 20,531 & 20,463 & 27,708 & \textbf{17,936}\\
    \emph{Queries made}& 433&484&484 & 467 & 469 & \textbf{403}
  \end{tabular}
  \end{center}
  \end{footnotesize}
  \vspace{-2mm}
  \caption{Ablation test on the features used to represent the RL state.}
  \vspace{-4mm}
  \label{table:ablation}
\end{table*}

  \begin{table}[t!]
  \begin{small}
  \begin{center}
  \begin{tabular}{l|ccc}
    &Empty query&Ungrounded&Low yield\\
    & results & participant(s) & from IE \\
    \hline
    \emph{Error cause}& 12&4&2
  \end{tabular}
  \end{center}
  \end{small}
  \vspace{-2mm}
  \caption{Error analysis on 18 queries that failed under the RL algorithm.}
  \vspace{-4mm}
  \label{table:policyerroranalysys}
\end{table}

We trained the RL Query Policy using the SARSA~\cite{sutton1998reinforcement} RL algorithm.  As the number of unique states is large, we used a linear approximation of the q-function.  Once the policy converged during training, we then fixed the linear estimate of the q-function and used this as a fixed policy for selecting queries.  
We trained the RL Query Policy on the separate set of 500 entity pairs, and 
evaluated it on the same data set of 289 participant pairs used to evaluate the baseline.
Table~\ref{table:reading-results} summaries the results of both the baseline and the RL Query Policy.  
The Query Policy resulted in a 25\% decrease in the number of queries that were run, leading to a 24\% drop in the number of papers that were read, while at the same time {\em increasing} the number of paths recovered by 3\%.
We tested the statistical significance of the difference in results between the baseline and RL policy by performing a bootstrap resampling test. Our hypotheses were that the policy reads fewer papers, makes fewer queries and finds more paths. The resulting estimated $p$-values for fewer papers and fewer queries was found to be near 0, and $< 0.003$ for finding more paths.
An ablation study of the state features found that features (f2) and (f5) had the largest impact on number of papers read; both model the history of the reading task (see the next section for details).  This highlights that the RL model is indeed learning to model the entire iterative process.


\section{Analysis}\label{sec:analysis}
\vspace{-2mm}
\paragraph{Feature Ablation Test:}

We performed an ablation test on the features that encode  the RL state. The results are summarized in Table \ref{table:ablation}. Similar to Section 5, we grouped the features into five different groups, and we measured the impact of removing one feature group at a time. Overall, the amount of paths found doesn't have a significant amount of variance, but the efficiency of the search (amount of papers read and number of queries made)  depends on several feature groups. 
For example, features (f1), (f2), and (f4) have a large effect on both the number of papers read and the number of queries made. 
Removing the feature (f5) actually reduces the number of papers read by approximately 2K with a minimal reduction in the number of paths found, which suggests that this task could benefit from feature selection. 


\paragraph{RL Policy Error Analysis:}

Lastly, we analyzed the execution trace of eighteen (20\% of the errors) of the searches that failed to find a path under RL. 
The results are summarized in Table \ref{table:policyerroranalysys}. 
The table shows that the main source of failures is receiving \emph{no results} from the information retrieval query, i.e., when the IR system returns zero documents for the chosen query. This is typically caused by over-constrained queries.
The second most common source of failures was {\em ungrounded participants}, i.e., when at least one of the selected participants that form the query could not be linked to our protein knowledge base. This is generally caused by mistakes in our NER sequence model, and also tends to yield no results from the IR component.
Finally, the \emph{low yield from IE} situation appears when the the information produced through machine reading in one iteration is scarce and adds no new components to the interaction graph, again resulting in a stop condition.
\section{Discussion and future work}

We introduced a framework for the focused reading of biomedical literature, which is necessary to handle the data overload that plagues even machine reading approaches.
  We have presented a generic focused reading algorithm, an intuitive, strong baseline algorithm that instantiates it,  and formulated an RL approach that learns how to efficiently query the paper repository that feeds the machine reading component. 
We showed that the RL-based focused reading is more efficient than the baseline (i.e., it reads 24\% fewer papers), while answering 7\% more queries.

There are many exciting directions in which to take this work.  First, more of the focused reading algorithm can be subject to RL, with the {\sc ChooseEndPoints} policy being the clear next candidate.  
Second, we can expand focused reading to efficiently search for multiple paths between $S$ and $D$.  
Finally, we will incorporate additional biological constraints (e.g., focus on pathways that exist in specific species) into the search itself. 

\section*{Acknowledgments}

This work was partially funded by the DARPA Big Mechanism
program under ARO contract W911NF-14-1-0395.

Dr. Mihai Surdeanu discloses a financial interest in Lum.ai. This interest has been disclosed to the University of Arizona Institutional Review Committee and is being managed in accordance with its conflict of interest policies.

\bibliography{emnlp2017}
\bibliographystyle{emnlp_natbib}

\end{document}